\RequirePackage{fix-cm}
\documentclass[12pt,onecolumn,final]{svjour3}                     
\smartqed  
\usepackage{graphicx}
 \usepackage{mathptmx}      
\usepackage{latexsym}
\usepackage{multirow}
\usepackage{multicol} 
\usepackage{amsfonts}
\usepackage{algorithmic}
\usepackage{array}
\usepackage[normalsize]{subfigure}
\usepackage{url}
\usepackage{color}
\usepackage{soul}

\usepackage{times}
\usepackage{helvet}
\usepackage{courier}
\usepackage{mathtools}
\usepackage{amsmath}
\usepackage{xcolor}
\usepackage{blindtext, graphicx}
\usepackage{amssymb}
\usepackage{gensymb}
\usepackage{latexsym}
\usepackage{multirow}
\usepackage{multicol} 
\usepackage{amsfonts}
\usepackage{algorithmic}
\usepackage{array}
\usepackage[normalsize]{subfigure}
\usepackage{url}
\usepackage{color}
\usepackage{soul}
\usepackage{cite}

\usepackage{rotating}
\usepackage[margin=1in]{geometry}
\usepackage{epstopdf}
\usepackage{nomencl}
\usepackage[numbers]{natbib}
 \journalname{Annals of Biomedical Engineering}
\begin{document}

\title{Fusing Continuous-valued Medical Labels using a Bayesian Model}


\author{Tingting~Zhu, Nic~Dunkley, Joachim~Behar, David~A.~Clifton, Gari~D.~Clifford}

\institute{The authors are with the Institute of Biomedical Engineering, Department of Engineering Science, University of Oxford, United Kingdom.
               G D Clifford is also with the departments of Biomedical Informatics and Biomedical Engineering at Emory University and Georgia Institute of Technology.\\
              Contact for Correspondence: Tingting Zhu\\
              \email{tingting.zhu@eng.ox.ac.uk}           
}

\date{Received: date / Accepted: date}

\maketitle

\clearpage

\begin{abstract}
\boldmath
With the rapid increase in volume of time series medical data available through wearable devices, there is a need to employ automated algorithms to label data. Examples of labels include interventions, changes in activity (e.g. sleep) and changes in physiology (e.g. arrhythmias). However, automated algorithms tend to be unreliable resulting in lower quality care. Expert annotations are scarce, expensive, and prone to significant inter- and intra-observer variance. To address these problems, a Bayesian Continuous-valued Label Aggregator(BCLA) is proposed to provide a reliable estimation of label aggregation while accurately infer the precision and bias of each algorithm.
\par
The BCLA was applied to QT interval (pro-arrhythmic indicator) estimation from the electrocardiogram using labels from the 2006 PhysioNet/Computing in Cardiology Challenge database. It was compared to the mean, median, and a previously proposed Expectation Maximization (EM) label aggregation approaches. While accurately predicting each labelling algorithm's bias and precision, the root-mean-square error of the BCLA was 11.78$\pm$0.63ms, significantly outperforming the best Challenge entry (15.37$\pm$2.13ms) as well as the EM, mean, and median voting strategies (14.76$\pm$0.52ms, 17.61$\pm$0.55ms, and 14.43$\pm$0.57ms respectively with $p<0.0001$). 
\par
The BCLA could therefore provide accurate estimation for medical continuous-valued label tasks in an unsupervised manner even when the ground truth is not available.

\end{abstract}


\section{Introduction}

\label{sec:ECG labelling issue}
With human annotation of data, significant intra- and inter-observer disagreements exist \cite{Warfield2008, OferDekel2009}. 
Expert labelling (or `reading' or `annotating') of medical data by physicians or clinicians often involves multiple over-reads, particularly when an individual is under-confident of the diagnosis. However, experts are scarce and expensive and can create significant delays in labelling or diagnoses. Although medical training includes periodic assessment of general competency, specific assessments for reading medical data are difficult to be performed regularly. This data processing pipeline is further complicated by the ambiguous definition of an `expert'. There is no empirical method for measuring level of expertise, even though label accuracy can vary greatly depending on the expert's experience. As a result, there exists a great deal of inter- and intra-expert variability among physicians depending on their experiences and level of training 
\cite{Warfield2008,OferDekel2009,Salerno2003,Metlay1997,Molinari2011,Valizadegan2012}.
\par
An effective probabilistic approach to aggregating expert labels which used an Expectation Maximization (EM) algorithm, was first proposed by Dawid and Skene \cite{Dawid1979}. They applied the EM algorithm to classify the unknown \textit{true} states of health (i.e. fit to undergo a general anaesthetic) of 45 patients given the decision made by five anaesthetists. Raykar\textit{ et al.} \cite{Raykar2010} extended this approach to measure the diameter of a suspicious lesion on a medical image using a regression model. Their assumption was that the discrepancies of the lesion diameter estimates from different expert annotators were Gaussian distributed and noisy versions of the actual \textit{true} diameter. The precision of each expert annotator and the underlying ground truth were jointly modelled in an iterative process using EM. Welinder and Perona \cite{Welinder2010} proposed a Bayesian EM framework for continuous-valued labels, which explicitly modelled the precision only of each annotator to account for their varying skill levels, without modelling the bias of annotators. A more specialised form of the Bayesian model of bias was proposed by Welinder\textit{ et al.} \cite{Welinder2010Wisdom} but for binary classification tasks. However, their model cannot account for more complex tasks such as the continuous-valued labelling.
\par
The methodology proposed in the work presented in this article improves on these prior algorithms \cite{Raykar2010,Welinder2010,Welinder2010Wisdom} by introducing the novelty of combining {\it continuous-valued annotations} to infer the underlying ground truth, while {\it jointly modelling the annotator's bias and precision} in an unified model using a Bayesian treatment.
\par
\begin{figure}[tp]
 \centering
    \includegraphics[trim=0cm 0cm 0.0cm 0cm, clip=true, width=1\textwidth]{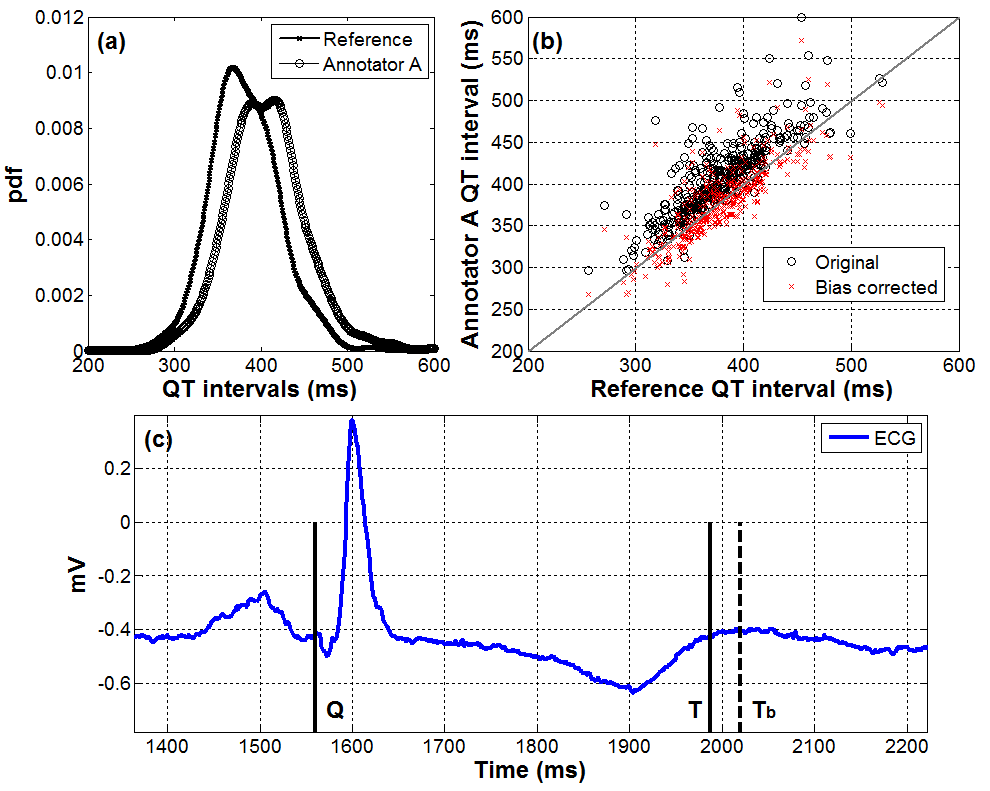}
\caption{An example of bias in the context of Electrocardiogram (ECG) QT interval labelling. (a) The probability density function of the QT intervals for the reference (supplied by the human experts) annotation and annotator A (such as an automated algorithm). A plot of QT intervals across different recordings: the diagonal (grey) line indicates a perfect match of QT intervals between the reference and annotator A; the `o' indicates the original QT intervals provided by annotator A; the `x' indicates the bias corrected QT intervals of annotator A, which fits closely to the diagonal line. (c) An example of bias that occurs in an ECG record for labelling QT interval. The reference QT interval on a single beat starts at the beginning of the Q wave and ends at the end of the T wave (denoted as $Q$ and $T$), and the biased trend from annotator A is demonstrated as $T_b$.}
 \label{fig:bias_example}
 \end{figure}

Aggregating annotations (i.e. fusing multiple annotations for each piece of data from annotators with varying levels of expertise) from human and/or automated algorithms may provide a more accurate ground truth and reduce annotator inter- and intra-variability. However, most annotators are likely to have some bias regardless of their expertise \cite{Welinder2010, zhu2014crowdlabel}. Bias is defined as the inverse of accuracy: It measures the average difference between the estimation and the \textit{true} value, and it is annotator dependent. An example of bias is demonstrated in Fig. \ref{fig:bias_example} in the context of Electrocardiogram labelling. Recently, Warby\textit{ et al.} \cite{warby2014sleep} studied how to combine non-expert annotator's labels of sleep spindle location, a special pattern in human electroencephalography, through fusing annotations provided by non-experts. In that work, although na\"ive majority vote was used to aggregate the labels of the locations, they demonstrated that non-expert annotations were comparable to those provided by the experts (i.e. the by-subject spindle density correlation was 0.815). Our proposed framework, in contrast, is a statistical approach that models the precision and bias of each annotator, which we hypothesise would provide a superior estimation of the ground truth as determined by a collection of experts.
\par
In contrast to previous works, this article proposes a Bayesian framework for aggregating multiple continuous-valued annotations in medical data labelling, which takes into account the precision and bias of the individual annotators. Moreover, we propose a generalised form which can be extended to incorporate contextual features of the physiological signal, so that we can adjust the weighting of each label based on the estimated bias and variance of the individual for different types of signal.
To our knowledge, the proposed model for estimating continuous-valued labels in an unsupervised manner is novel in the medical domain. 

\section{Materials and Methods}
\subsection{Bayesian Continuous-valued Label Aggregator (BCLA)} 

\begin{figure}[tp]
  \centering
 	    \includegraphics[trim=0cm 0cm 0.0cm 0cm, clip=true, width=0.65\textwidth]{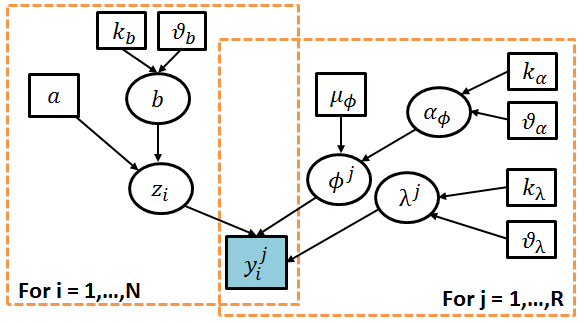}
  \caption{Graphical representation of the BCLA model: ${y}^{j}_{i}$ corresponds to the annotation provided by the $j$th annotator for the $i$th record, and it is modelled by the $z_{i}$ (the unknown underlying ground truth), the ${\bf\phi}^{j}$ (bias), and the ${\lambda}^{j}$ (precision). Furthermore, $z_{i}$ is drawn from a Gaussian distribution with parameters mean $a$ and variance $1/b$, where $a$ can be a function of feature vector ${\bf x}_{i}$. ${\phi}^{j}$ is modelled from a Gaussian distribution with mean $\mu_\phi$ and variance $1/\alpha_\phi$. The $b$, ${{\lambda}^{j}}$, and $\alpha_\phi$ are drawn from a Gamma distribution (denoted as $\Gamma$) with parameters $k_{b}$, $\mathrm{\vartheta_{b}}$, $k_{\lambda}$, $\mathrm{\vartheta_{\lambda}}$, and $k_{\alpha}$, $\mathrm{\vartheta_{\alpha}}$ respectively. }
  \label{fig:BCLA_model}
 \end{figure}

Suppose that there are $N$ records of physiological time series data labelled by $R$ annotators. Let ${\bf D}=\left[{\bf x}^\intercal_{i}, y^{j=1}_{i},\cdots, y^{j=R}_{i}\right]^{N}_{i=1}$, where ${\bf x}_{i}$ is a column feature vector for the $i$th record containing $d$ features (i.e. the design matrix, $\bf X=[{\bf x}^\intercal_{1},..., {\bf x}^\intercal_{N}]$), ${y}^{j}_{i}$ corresponds to the annotation provided by the $j$th annotator for the $i$th record, and ${z}_{i}$ represents the unknown underlying ground truth (the \textit{true} time or duration of an event for example). The graphical representation of the proposed approach -- the Bayesian Continuous-valued Label Aggregator (BCLA) -- is shown in Fig. \ref{fig:BCLA_model}.
\par
In this model, it is assumed that ${y}^{j}_{i}$ was a noisy version of $z_{i}$, with a Gaussian distribution ${\mathcal{N}(y^{j}_{i}\mid z_{i},({\bf \sigma}^{j})^{2})}$\footnote{The motivation for this model comes from the Central Limit Theorem. Given the assumption that the annotators are independent and identically distributed, their labels will converge to a Gaussian distribution. In the absence of prior knowledge, this assumption allows for a robust and generalizable model for the given data.}. Here ${\bf \sigma}^{j}$ is the standard deviation of the $j$th annotator and represents his variance in annotation around $z_{i}$. Furthermore, the bias of each annotator can be modelled as an additional term, ${\bf\phi}^{j}$. The probability of estimating  ${y}^{j}_{i}$ can be written as: 
\begin{equation}
\label{eq:yijPr}
\mathrm{P}[y^{j}_{i}\mid {z}_{i},({\bf \sigma}^{j})^{2}]=\mathcal{N}(y^{j}_{i}\mid {z}_{i}+{\phi}^{j},1/{\lambda}^{j}). 
\end{equation}
where ${({\bf \sigma}^{j})^{2}}$ is replaced with $1/{\lambda}^{j}$. ${\lambda}^{j}$ is the precision of the $j$th annotator, defined as the estimated inverse-variance of annotator $j$. Note that ${\lambda}^{j}$ and ${\phi}^{j}$ are considered to be constants for the $j$th annotator, i.e. all annotators are assumed to have consistent but usually different performances throughout records. Furthermore, it is assumed that the probability of a given bias of annotator $j$, ${\phi}^{j}$, is drawn from a Gaussian distribution with mean $\mu_\phi$ and variance $1/\alpha_\phi$, is given by:
\begin{equation}
\label{eqn:phiPr}
\mathrm{P}[\phi^{j}\mid \mu_\phi,\alpha_\phi]=\mathcal{N}(\phi^{j}\mid \mu_\phi,1/\alpha_\phi).
\end{equation}
Although the biases of the annotators might be derived from other distributions, they are likely to be data set dependent. In the absence of any knowledge of the underlying distribution of biases, they are assumed to be drawn from a Gaussian distribution.
Furthermore, the ground truth, $z_{i}$, can be assumed to be drawn from a Gaussian distribution with mean ${a}$ and variance ${1/b}$. The probability of $z_{i}$ is defined as follows:
\begin{equation}
\label{eqn:yiPr}
\mathrm{P}[z_{i}\mid a,b]=\mathcal{N}(z_{i}\mid a,1/b),
\end{equation}
where ${a}$ can be expressed as a linear regression function $f( \bf{w},\bf x)$ with an intercept, and $ {\bf w}$ being the coefficients of the regression \cite{Zhu2014, Raykar2010}. The intercept models the overall offset predicted in the regression, which is different from the annotator specific bias in the proposed model. 
Under the assumption that records are independent, the likelihood of the parameter ${\bf \theta} = \lbrace{\bf w},{\bf \lambda},{\bf\phi}, {\alpha_\phi}, b, z_{i} \rbrace$ for a given data set ${\bf D}$ can be formulated as:
\begin{equation}
\mathrm{P}[{\bf D} \mid{\bf \theta}]= \prod^{N}_{i=1}\mathrm{P}[y^{1}_{i},\cdots,y^{R}_{i}\mid {\bf x}_{i},{\bf \theta}].
\end{equation}
It is assumed that $y^{1}_{i},\cdots, y^{R}_{i}$ are conditionally independent given the feature ${\bf x}_{i}$ (i.e. each annotator works independently to provide annotations). This may or may not be necessarily true, especially in cases where the annotations are generated by algorithms, some of which may be variations of the same approach. Nevertheless, this assumption was made to simplify the model and subsequent derivation of the likelihood. The likelihood of the parameter ${\bf \theta}$ for a given data set ${\bf D}$ can be written using the Bayes' theorem as (see detailed description in Fig. \ref{fig:BCLA_model}):
\begin{eqnarray}
\label{eqn:theta}
\mathrm{P}[{\bf \theta}\mid{\bf D}]\propto \mathrm{P}[{\bf D}\mid{\bf \theta}] \cdot \mathrm{P}[{\bf \theta}] \nonumber \\
= \Gamma({\alpha_\phi}\mid k_{\alpha},\vartheta_{\alpha}) \lbrack \prod^{R}_{j=1}\mathcal{N}({\phi}^{j}\mid  \mu_\phi,1/\alpha_\phi)\Gamma({\lambda}^{j}\mid k_\lambda,\vartheta_\lambda)\rbrack  \nonumber \\
\Gamma({b}\mid k_{b},\vartheta_{b}) \lbrack \prod^{N}_{i=1}\mathcal{N}(z_{i}\mid  a,1/b) \prod^{R}_{j=1}\mathcal{N}(y^{j}_{i}\mid  z_{i}+{\phi}^{j},1/ {\lambda}^{j})\rbrack. 
\end{eqnarray}

where $\Gamma$ denotes a Gamma distribution and can be defined as ${\Gamma(z\mid k, \vartheta)=\frac{1}{\Gamma(k)\vartheta^{k}}z^{k-1}exp(-\frac{z}{\vartheta})}$, where $k$ is the shape of the distribution and $ \vartheta$ is the scale of the distribution. Gamma distribution is commonly used to model positive continuous values. It is therefore assumed that precision values, such as $b$, ${\lambda}^{j}$, and $\alpha_\phi$ were drawn from a Gamma distribution, with parameters $k_{b}$, $\mathrm{\vartheta_{b}}$, $k_{\lambda}$, $\mathrm{\vartheta_{\lambda}}$, and $k_{\alpha}$, $\mathrm{\vartheta_{\alpha}}$ respectively.

\subsection{The Maximum a posteriori approach}
The estimation of ${\bf \theta}$ can be solved using the maximum a posteriori (MAP) approach, which maximises the log-likelihood of the parameters, i.e. ${\mathrm{arg}\underset{\mathrm{\bf \theta}}{\mathrm{max}} \lbrace \log \mathrm{P}[{\bf \theta}\mid{\bf D} ]\rbrace}$. The log-likelihood can be rewritten as:
 \begin{eqnarray}
 \log \mathrm{P}[{\bf \theta} \mid{\bf D}] = -\frac{1}{2}\sum_{i=1}^{N} \sum_{j=1}^{R}[\log (\frac{2\pi} {{\lambda}^{j}})+(y_{i}^{j}-{\phi}^{j}- z_{i})^{2}{\lambda}^{j}]\nonumber \\
  - \frac{1}{2}\sum_{j=1}^{R}[\log (\frac{2\pi}{\alpha_\phi})+{({\phi}^{j}-\mu_\phi})^{2}\alpha_\phi] \nonumber \\
 - \frac{1}{2}\sum_{i=1}^{N}[\log (\frac{2\pi} {b})+(z_{i}-{\bf x}_{i}^{^\intercal}{\bf w})^{2}{b}] \nonumber \\
 + [(k_\lambda-1)\log {\lambda}^{j}- \log(\Gamma(k_\lambda)\vartheta_\lambda^{(k_{\lambda})}- \frac { {\lambda}^{j}}{\vartheta_\lambda}]\nonumber \\
 + [(k_{\alpha}-1)\log {\alpha_\phi}- \log(\Gamma(k_{_\alpha})\vartheta_{\alpha}^{(k_{\alpha})})- \frac {\alpha_\phi}{\vartheta_{\alpha}}]\nonumber \\
 + [(k_{b}-1)\log {b}- \log(\Gamma(k_{_b})\vartheta_{b}^{(k_{b})})- \frac {b}{\vartheta_{b}}].
 \end{eqnarray}
The parameters in ${\bf \theta}$ can be derived by equating the gradient of the log-likelihood to zero respectively as follows:
\begin{equation}
\label{eqn:lambda_j}
\frac{1}{{\bf \lambda}^{j}} = \frac{1}{N+2(k_\lambda-1)}[\sum_{i=1}^{N}(y_{i}^{j}-{\phi}^{j}- z_{i})^{2}+\frac{2}{\vartheta_\lambda}].
\end{equation}
\begin{equation}
\label{eqn:w_j}
{\bf w} = {(\sum_{i=1}^{N}{\bf x}_{i}{\bf x}_{i}^{^\intercal})}^{-1}\sum_{i=1}^{N}{\bf x}_{i} z_{i}.
\end{equation}
\begin{equation}
\label{eqn:phi_j}
{\bf\phi}^{j} = \frac{1}{N+\frac{\alpha_\phi}{\lambda^{j}}}[\sum_{i=1}^{N}(y_{i}^{j}-z_{i})+\mu_\phi (\frac{\alpha_\phi}{\lambda^j})].
\end{equation}
\begin{equation}
\label{eqn:alpha}
\frac{1}{\alpha_\phi}= \frac{1}{R+2(k_{\alpha}-1)}[\sum_{j=1}^{R}{(\phi ^{j}-\mu_\phi)}^{2}+\frac{2}{\vartheta_{\alpha}}].
\end{equation}
\begin{equation}
\label{eqn:yi}
z_{i}=  \frac{ \sum_{j=1}^{R}[(y_{i}^{j}-{\phi}^{j}){\lambda}^{j}]+({\bf x}^\intercal_{i}{\bf w}){b}}{\sum_{j=1}^{R}{\lambda}^{j}+{b}}.
\end{equation}
\begin{equation}
\label{eqn:b}
\frac{1}{b}=   \frac{1}{N+2(k_{b}-1)} [\sum_{i=1}^{N}{(z_{i}-{\bf x}^\intercal_{i}{\bf w})^{2}}+\frac{2}{\vartheta_{b}}].
\end{equation}

This MAP problem can be solved using the EM algorithm in a two-step iterative process:\\
i) The E-step estimates the expected {\it true} annotations for all records, ${\hat{\bf z}}$, as a weighted sum of the provided annotations, 
and can be estimated using equation (\ref{eqn:yi}).\\
ii) The M-step is based on the current estimation of $\hat{\bf z}$ and given the data set $\bf D$. The model parameters, ${\bf {w}}$, $\phi$, $\alpha_\phi$, $b$, and $\bf \lambda$ can be updated using equations (\ref{eqn:w_j}), (\ref{eqn:phi_j}), (\ref{eqn:alpha}), (\ref{eqn:b}), and (\ref{eqn:lambda_j}) accordingly in a sequential order until convergence, which is now described.

\subsection{Convergence criteria for the MAP-EM approach}

When solving a MAP-EM algorithm one may encounter a convergence issue, particularly when estimating a large number of parameters. The estimation of the precision may approach to infinity because the inferred annotations favour the annotator with the highest precision in each EM update step while maximising the likelihood. Instead of incorporating an additional parameter for the regularisation penalty that increases the complexity of the mode, the generalized extreme value distribution (GEVD) can be used to model the maxima of the precision distribution, denoted as $\lambda_m$, in order to restrict the upper bound of the precision values and guarantee a convergence in the MAP algorithm. The probability density function of the GEVD for $\lambda_m$ can be expressed as: 
 \begin{eqnarray}
\label{eqn:GEVD}
 \mathrm{P}(\lambda_m \mid k,\mu,\vartheta) = \exp\lbrace -[1+k\frac{(\lambda_m-\mu)}{\vartheta}]^{-\frac{1}{k}}\rbrace 
 \frac{1}{\vartheta}[1+k\frac{(\lambda_m-\mu)}{\vartheta}]^{(-1-\frac{1}{k})},
 \end{eqnarray}
where $k$ is the shape parameter, $\vartheta$ is the scale parameter, and $\mu$ is the location parameter. These parameters can be derived by fitting a GEVD to the maximum values drawn randomly from the {\it prior} distribution of the precision, $\Gamma({\lambda}\mid k_\lambda,\vartheta_\lambda)$. An upper bound of the maximum precision value can then be obtained by estimating the $99th$ quantile of the inverse cumulative distribution function of the GEVD.

\subsection{Data description}

The electrocardiogram (ECG) is a standard and powerful tool for assessing cardiovascular health as many detrimental heart conditions manifest as abnormalities in the ECG. The QT interval is one particular measure of ECG morphology, and refers to the elapsed time between the onset of ventricular depolarisation (the QRS complex) and the T wave offset (ventricular repolarisation) \cite{CliffordBook2006}. Accurate measurement of the QT interval  is essential since abnormal intervals indicate a potentially serious but treatable condition, and can be a contraindication for the use of drugs or other interventions \cite{ICH2005}. 
Viskin\textit{ et al.} \cite{Viskin2005} presented the ECGs recorded from two patients with long QT syndrome (LQTS) and from two healthy females to 902 physicians (25 QT experts who had published on the subject, 106 arrhythmia specialists, 329 cardiologists, and 442 noncardiologists) from 12 countries. No other details were given on actual training or intrinsic accuracy of these annotators. For patients with LQTS, 80\% of arrhythmia specialists calculated the QTc (the heart rate corrected QT interval) correctly but only 50\% of cardiologists and 40\% of noncardiologists did so. 
In the context of QT annotation where baseline wander is frequent, it was observed that a few annotators consistently over- or under-estimated the QT interval \cite{zhu2014crowdlabel}. Other studies have reported significant intra- and inter-observer variability in QT annotations, ranging from 10 to 30ms \cite{Ehlert1992,Christov2006}. 
It is important to note that experts or non-experts with different levels of training or expertise can have significantly different biases. Na\"ive approaches to aggregate labels from a group of annotators of unknown expertises could therefore lead to poor results. However, annotators' biases are rarely taken into account when aggregating different labels or opinions in medical labelling tasks. 
\par
We hypothesise that incorporating an accurate estimation of each annotator's bias into a model for fusing annotations (as described in sections 2.1 to 2.3) will result in an improved estimate of the ground truth. In order to test this hypothesis we have used two data sets: one simulated data set to ensure an absolute ground truth is available; and one real data set of QT intervals. Although we have chosen to use QT interval data, because of the availability of the numerous annotations, the method we present is more general and can be applied to other continuous-valued annotations..

\subsubsection{Simulated data set}

\begin{figure}[tp]
\begin{center}
\includegraphics[scale=0.8]{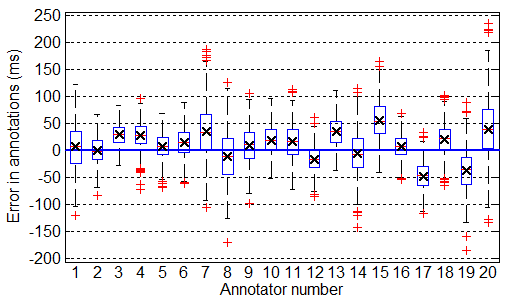}
\end{center}
\caption{The box plot of the error between the generated and \textit{true} annotations for each of the 20 simulated annotator. The black `x' indicates the bias of each annotators. The span of each box represents the precision of the annotations (rather than the interquartile range) over all annotations for each annotator.}
\label{fig:simulated annotators}
\end{figure}

To test the reliability of the BCLA as a generative model, a simulated data set was created: a total of 548 simulated records were generated, each has 20 independent annotator, thus providing a total of 10,960 annotations (see Fig.~\ref{fig:simulated annotators}). The simulated data set considered that annotators have precision values, $\bf \lambda$ (i.e. $1/\sqrt{\bf \sigma}$), which were drawn from $\Gamma(4,0.0003)$, with assumption that the annotations provided by the best performing annotator is $\pm15$ms away from the ground truth. Annotators' biases were drawn from $\mathcal{N}(10,25)$, a Gaussian distribution with 10ms mean and a standard deviation ($1/\sqrt{\alpha_\phi}$) of 25ms. The \textit{true} annotation for each record was drawn from $\mathcal{N}(400,40)$, a Gaussian distribution with a mean, $a$, of 400ms with a standard deviation ($1/\sqrt{b}$) of 40ms. 
In addition, it was assumed that $\alpha_\phi$ was drawn from $\Gamma(3, 0.0005)$, ensuring the mean standard deviation where the biases drawn from is $25$ms. The $b$ was drawn from $\Gamma(3, 0.0002)$, ensuring the mean standard deviation where the \textit{true} annotations drawn from is $40$ms. The generated 10,960 annotations were then fed into the BCLA model to evaluate its accuracy in estimating the \textit{true} annotation in an unsupervised manner as well as predicting the bias and precision of each annotator. 

\subsubsection{Real data set}
The data were drawn from the QT interval annotations generated by participants in the 2006 PhysioNet/Computing in Cardiology (PCinC) Challenge \cite{Moody2006} for labelling QT intervals with reference to Lead II in each of the 548 recordings in the Physikalisch-Technische Bundesanstalt Diagnostic ECG Database (PTBDB) \cite{Bousseljot1995}. The records were from 290 subjects (209 men with mean age of 55.5 years and 81 women with mean age of 61.6 years), in which 20\% of the subjects were healthy controls. An example of QT interval is demonstrated in Fig.~\ref{fig:bias_example}(c). The PTBDB database contained records of patients with a variety of ECG morphologies having different QT intervals ranging from 256 to 529 ms. The diagnostic classifications of ECG morphologies mainly included myocardial infarction, heart failure, bundle branch block, and dysrhythmia as stated in Bousseljot and Kreiseler \cite{Bousseljot1995}.
\par
There were two main categories of annotations: manual and automated (see Table~\ref{table:data description of 5-second segment}). A total of 38,621 annotations were collected and were divided into three divisions: 20 human annotators in Division 1, 48 closed source automated algorithms in Division 2, and 21 open source automated algorithms in Division 3. Division 4 was further created here so as to combine all automated algorithms from Division 2 and 3 in order to provide a larger data set and allow a better estimation of automated QT intervals. The number of annotators per division and averaged number of annotations per record are listed in Table~\ref{table:data description of 5-second segment}. The overall percentage of the annotators in each division with complete annotations (i.e. annotations on all 548 recordings) was: 55\% in Division 1, 40\% in Division 2, 43\% in Division 3, and  45\% in Division 4. The competition score for each entry was calculated from the root mean square error (RMSE) between the submitted and the reference QT intervals. The reference annotations were generated from Division 1's entries using a maximum of 15 participants by taking the ``median self-centering approach" as reported by the competition organisers as detailed in \cite{Willems1985}. The best-performing score for each division is also listed in Table~\ref{table:data description of 5-second segment}. Furthermore, the majority of the QT annotations of each 2-minute record occurred within the first 5 seconds of the ECG recordings. The best scores in the first 5-second segment were similar to those of the 2-minute segment (denoted by $\star$ in Table~\ref{table:data description of 5-second segment}). To reduce any possible inter-beat variations, only the annotations within the first 5-second segment of each record were chosen to ensure that all annotators had approximately labelled the same region of a record with similar QT morphologies. Therefore, the motivation for choosing the first 5-second segment of each record was to consider a short segment where the QT interval is not changing dramatically (with respect to a particular beat an annotator chose), while retaining the highest number of annotations. Those that fell outside this segment were considered to be missing information and discarded in the process of the QT estimation.
\par
As the manual entry (i.e. Division 1) was used to generate the reference annotations, we therefore focused on the analysis of the automated entry (i.e. Division 2, 3, and 4). In terms of parameter setting (see Table~\ref{table:BCLA_parameters}), annotator specific precision was drawn from $\Gamma(k_\lambda,\vartheta_{\lambda})$, with assumption that the annotations provided by the best performing algorithm is $\pm5$ms away from the reference. Annotators' biases were considered to be drawn from $\mathcal{N}(\mu_\phi,1/\sqrt{\alpha_\phi})$, and $\alpha_\phi$ was modelled by $\Gamma(k_\alpha,\vartheta_{\alpha})$, assuming that the automated annotations tend over-estimate manual annotations as described in previous studies \cite{Hughes2006, Couderc2011, Worster2014}. The \textit{true} QT interval for each record was assumed to be drawn from $\mathcal{N}(a,1/\sqrt{b})$, where $b$ was modelled by $\Gamma(k_b,\vartheta_b)$ \cite{Malik2002, Goldenberg2006, CliffordBook2006}. Instead of assuming the mean (i.e. $a$) of the underlying ground truth to be a fixed scalar, we updated it using a linear regression function, $f( \bf{w},\bf x)$, where the coefficients, $\bf w$, were estimated using equation (\ref{eqn:w_j}). An intercept was included in $f( \bf{w},\bf x)$ to model the overall offset predicted in $f$, and no particular features were considered in this case (i.e. ${x}_i =1$) as we were solely interested in the performance of the model. 

\subsection{Methodology of validation and comparison}
The BCLA inferred precision of individual algorithms was compared with those estimated using the EM algorithm proposed by Raykar\textit{ et al.} \cite{Raykar2010} (denoted as EM-R) as it served as one of the benchmarking algorithms. Furthermore, the mean and standard deviation ($\mu\pm \sigma_{\mu}$ms) of 100 bootstrapped (i.e. random sampling with replacement) samples across records from the BCLA model were compared with the best algorithm (i.e. the algorithm with highest precision after correction of the bias offset), EM-R, and the traditional na\"ive mean and median voting approaches in both simulated and real data sets. The mean absolute error (MAE) of the annotations was also calculated as it provides interpretation of the difference between the estimated and the reference annotations (with a resolution of 1ms). A two-sided Wilcoxon rank sum test ($p<0.0001$) was applied to the 100 bootstrapped RMSEs and MAEs, to provide a comparison for the BCLA and EM-R versus other methodologies. In assessing the performance of the BCLA as a function of the number of annotators, a random number of annotators was selected 100 times. This was repeated with the annotator numbers varied from three to the maximum number of annotators in the division. The minimum number of annotators was chosen to be three to allow for obtaining results from the median voting approach. The $\mu\pm \sigma_{\mu}$ms of the RMSE of the BCLA, the EM-R, the mean, and the median were calculated and compared.

\section{Results}

The convergence of the BCLA model is guaranteed by providing a threshold using the GEVD as a stopping criteria (see Eqn~(\ref{eqn:GEVD})). In the real data set, the upper bound of the precision derived from the GEVD was 0.04, which was based on the assumption that the best performing annotator is $\pm 5$ms away from the reference. The number of iteration is dependent on the number of records and the number of annotations. To illustrate the practical utility of our model, it took 7.55 seconds for the BCLA to perform 5,000 iterations when considering a total of 20,712 annotations (Division 2) using MATLAB R2011a on a 2.2GHz Intel(R) i7-2670QM processor. Approximately 2,500 iterations were required to stabilise all the parameters.

\subsection{Simulated data set}
\begin{figure}[tp]
\begin{center}
\includegraphics[scale=0.80]{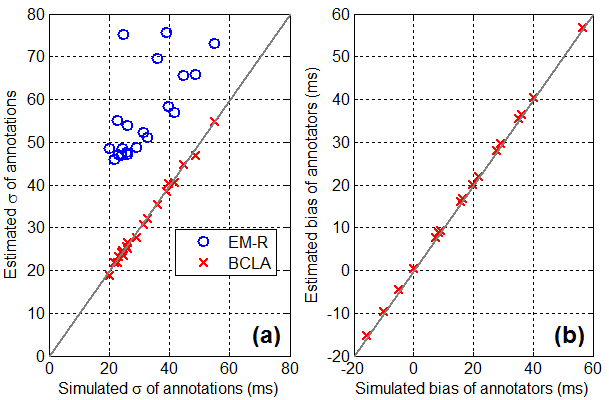}
\end{center}
\caption{A comparison of the simulated and inferred $\sigma$ in (a) and bias in (b) of each annotator in the simulated data set. The precision can be estimated by taking $1/(\sigma)^{2}$. The diagonal (grey) line indicates a perfect match between simulated and estimated results. Note that EM-R significantly over-estimates the $\sigma$ in all simulations.} 
\label{fig:simulation}
\end{figure}

Fig.~\ref{fig:simulation}(a) shows an example of the inferred results estimated using the EM-R and the BCLA. As the EM-R algorithm modelled jointly the precision (i.e. $1/(\sigma)^{2}$) of each annotator and the noise of underlying ground truth, its estimated $\sigma$ cannot represent the real precision of each annotator. Furthermore, EM-R algorithm does not consider the bias of each annotator, and we observe that its estimated values of $\sigma$ were well above the line of identity, indicating a consistent over-estimation. 
In contrast, the BCLA inferred results of $\sigma$ lie closely to the line of identity in the plot, indicating that the BCLA model can provide a reliable estimation of the \textit{true} precision in the simulated results. In addition to precision, the BCLA modelled the bias of each annotator and the results are provided in Fig.~\ref{fig:simulation}(b): the estimated biases are very close to the {\it true} biases. Although not all the estimated precisions and biases of each annotator were identical to the simulated values, the BCLA model inferred annotations without any prior knowledge of who the best annotator was in an unsupervised manner. 
\par
In order to compare the accuracy of the inferred labels using the BCLA model, the simulated 548 annotations were bootstrapped 100 times. Each time a RMSE and MAE were generated and compared to the best annotator, mean, EM-R, and median voting strategies. The results are shown in Table~\ref{table:BCLA_sim_comparison}. The RMSE and MAE results show that BCLA inferred labels significantly outperformed the mean, median, EM-R, and best annotator when compared with the simulated \textit{true} annotations.

\subsection{Real data set}

\begin{figure*}[tp]
\begin{center}
\includegraphics[scale=0.50]{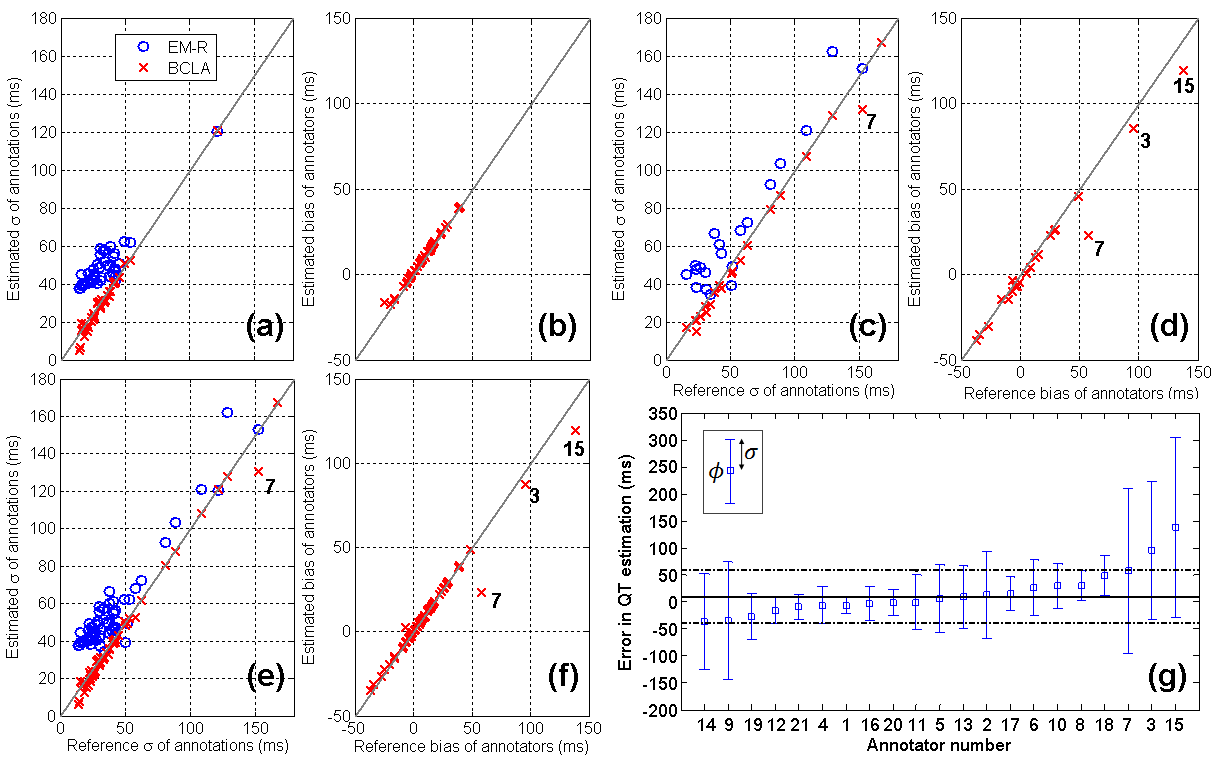}
\end{center}
\caption{A comparison of the 2006 PCinC Challenge reference and inferred $\sigma$ and bias of each annotator using the reference provided for division 2 in (a) and (b), division 3 in (c) and (d), and division 4 in (e) and (f) respectively. The precision can be estimated by taking $1/(\sigma)^{2}$. The leading diagonal line of each plot indicates a perfect matched between the Challenge reference and the estimated results. The mean (i.e. bias), $\phi$, and $\sigma$ of the difference in annotations for Division 3 are shown in (g). The annotators were ranked based on their bias values. The solid line indicates the mean of the biases whereas the dotted lines indicate 1.96$\sigma$ of the mean assumed in the BCLA. Note the annotator 3, 7, and 15 are labelled in the corresponding plots.}
\label{fig:PCinC_results_bias_prec}
\end{figure*}

Fig.~\ref{fig:PCinC_results_bias_prec} (a) to (f) show the inferred precision and bias results estimated using EM-R and BCLA for different automated divisions. As mentioned previously, the EM-R algorithm does not directly model the precision (i.e. $1/(\sigma)^{2}$) of each annotator; its estimated $\sigma$ of each annotator produces an offset from the values provided by the reference annotations. In contrast, the BCLA inferred $\sigma$ results lie much closer to the line of identity in the Fig.~\ref{fig:PCinC_results_bias_prec} (a), (c), and (e), indicating that the BCLA model can provide a reliable estimation of the \textit{true} precision of each annotator. In addition, the BCLA modelled the bias of each annotator accurately (see Fig.~\ref{fig:PCinC_results_bias_prec} (b), (d), and (f)). 
Although automated annotator 3 and 15 were predicted by the BCLA to have lower bias values than those provided by the reference, they are considered to be outliers due to the assumption made in our model: annotators' biases were drawn from a Gaussian distribution with 10ms mean and 25ms standard deviation. As Fig.~\ref{fig:PCinC_results_bias_prec} (g) shows, the biases of annotator 3 and 15 lie outside the 95\% of the area  (i.e. $\pm$1.96$\sigma$ of the mean under the normal distribution) predicted by the BCLA. In the case of annotator 7, its precision was underestimated (see  Fig.~\ref{fig:PCinC_results_bias_prec}(c) and (e)), which also affected the BCLA's estimation of its bias value. It was observed  that only 3.47\% of records were annotated by annotator 7, making it harder for the BCLA to provide a reliable estimation of its precision and bias values. 
In the evaluation of the inferred labels, the 548 records were bootstrapped 100 times, the RMSEs and MAEs of the BCLA model were generated and compared to the best annotator, mean, EM-R, and median voting approaches for the given reference. The results are displayed in Table~\ref{table:BCLA_qt_comparison}:

for Division 2 using 48 algorithms, the BCLA achieved a RMSE of 12.57$\pm$0.67ms, which significantly outperformed other approaches and provides an improvement of 16.48\% over the next best approach (EM-R with RMSE of 15.05$\pm$0.49ms); in the closed source entry Division 3 using 21 algorithms, the BCLA again exhibited a superior performance over the other methods with a RMSE of 13.90$\pm$0.84, and a 19.48\% improved error rate over the next best method (RMSE of 17.25$\pm$2.33ms). When considering all automated entries (Division 4), the BCLA provided an even more accurate performance than on the other two data sets (Division 2 and 3) as well as over other methods tested with a RMSE of 11.78$\pm$0.63ms. 

\begin{figure}[tp]
\begin{center}
\includegraphics[scale=0.85]{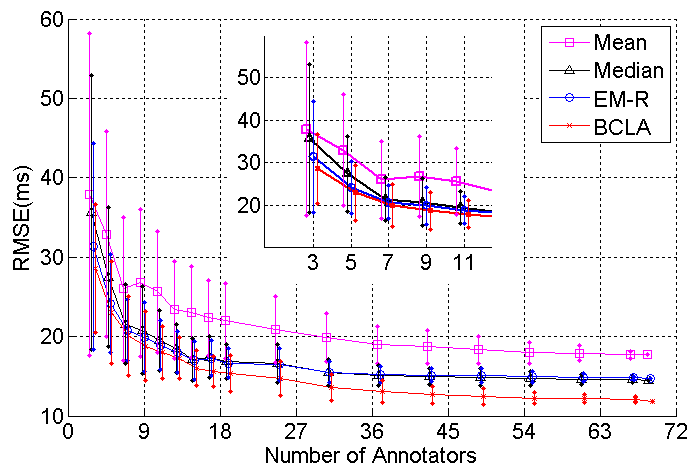}
\end{center}
\caption{The mean and standard deviation of the RMSE results as a function of the number of annotators for Division 4 when using the BCLA, EM-R, median, and mean voting approaches. Inset: A close-up of the RMSE results when using 11 annotators or less. }
\label{fig:PCinC_results_all}
\end{figure}

A further evaluation of the accuracies in terms of RMSE were made as a function of the number of annotators (see Fig.~\ref{fig:PCinC_results_all}). The results were generated by sub-sampling annotators with no replacement 100 times. The benchmarking algorithm, EM-R outperformed mean and median approaches initially but then underperformed when compared to the median approach after 43 algorithms are used. The BCLA model outperformed the other methods being tested with any number of annotators considered. 
In practice, it is rare to have more than three to five independent algorithms for estimating a label or predicting an event. In the case where only three automated algorithms were randomly selected, the BCLA had on average 9.02\%, 19.82\%, and 24.56\% improvement over the EM-R, median and mean voting approaches respectively. 

Although the lowest BCLA RMSE (11.78$\pm$0.63ms) in the automated entry is larger than the best-performing human annotator in the Challenge (RMSE = 6.65ms), there were only two other human annotators who achieved a score below 10ms. Furthermore, as the annotations of automated algorithms were independently determined from the reference, whereas the reference includes the best human annotators, it is unsurprising that a combination of the automated algorithms would have worse performance.

\section{Discussion}

In this article, a novel model, Bayesian Continuous-valued Label Aggregator, was proposed to infer the ground truth of continuous-valued labels where accurate and consistent expert annotations are not available. As a proof-of-concept, the BCLA was applied to the QT interval estimation from the ECG using labels from the 2006 PhysioNet/Computing in Cardiology Challenge database, and it was compared to the mean, median, and a previously proposed Expectation Maximization label aggregation methods (i.e. EM-R). While accurately predicting each labelling participant’s bias and precision, the root-mean-square error of the BCLA algorithm was significantly outperformed the best Challenge entry as well as the EM-R, mean, and median voting strategies. There are two key contributions in our approach: i) the BCLA provides an estimation of {\it continuous-valued annotations} which is valuable for time-series related data as well as duration of events for physiological data; ii) It introduces a unified framework for combining {\it continuous-valued annotations} to infer the underlying ground truth, while {\it jointly modelling annotators' biases and precisions}. The BCLA operates in an unsupervised Bayesian learning framework; no reference data were used to train the model parameters and a separate training and validation test sets were not required. 
Combining more experienced annotators would therefore provide a better estimation of the inferred ground truth. Importantly though, the BCLA does {\it guarantee a performance better than the best annotator without any prior knowledge of who or what is the best annotator}.
\par
Novel contextual features were introduced in our previous study \cite{Zhu2014} which allowed an algorithm to learn how varying physiological and noise conditions affect each annotator's ability to accurately label medical data. The inferred result was shown to provide an improved `gold standard' for medical annotation tasks even when the ground truth is not available. As the next step, if we incorporate the context into the weighting of annotators, the BCLA is expected to have an even larger impact for noisy data sets or annotators with a variety of specialisations or skill levels. The current model assumed consistent performance of each annotator throughout the records: i.e. that is his/her performance is time-invariant. Although this might not be true over an extended period of time where an annotator’s performance might improve through learning, or their performance might drop due to inattention or fatigue, the nature of the data sets being considered in this work are such that we can assume that performance across records is approximately consistent for each annotator. Future work will include modelling the performance of each annotator varying across records and through time to provide a more reliable estimation of the aggregated ground truth for data sets in which intra-annotator performance is highly variant.
\par
Our model of the annotators currently does not factor in the possible dependency/correlation between individual annotators, which might not be the case for automated algorithms. Incorporating a correlation measure into the annotator's model could possibly allow for a better aggregation of the inferred ground truth. Annotators who are considered to be anomalous (i.e. highly correlated but have large variances and biases) should be penalised with lower weights; expert annotators (i.e. highly correlated but have small variances and biases) should be favourably voted in the model. Finally, combining annotations derived from reliable experts using the BCLA model could potentially lead to improved training for supervised labelling approaches.

\begin{acknowledgements}
TZ acknowledges the support of the RCUK Digital Economy Programme grant number EP/G036861/1 and an ARM Scholarship in Sustainable Healthcare Technology through Kellogg College. ND was supported be Cerner Corporation and the UK EPSRC. JB was supported by the UK EPSRC, the Balliol French Anderson Scholarship Fund, and MindChild Medical Inc. DAC is supported by the Royal Academy of Engineering and Balliol College. The final publication is available at Springer via http://dx.doi.org/10.1007/s10439-015-1344-1.
\end{acknowledgements}

\bibliographystyle{spmpsci}      
\bibliography{Ref}

\newpage

\clearpage

\clearpage

\clearpage

\begin{table}[tp]
\caption{Performance by competition entrants on the first 5-second ECG segment for each division of the 2006 PCinC Challenge.}
\centering  
\resizebox{15cm}{!} {
\begin{tabular}
{>{\centering\arraybackslash}p{3cm}>{\centering\arraybackslash}p{3.7cm}>{\centering\arraybackslash}p{1.7cm}>{\centering\arraybackslash}p{1.7cm}>{\centering\arraybackslash}p{1.7cm}}
\hline\noalign{\smallskip}                
& \normalsize{\textbf{Manual annotators}}  & \multicolumn{3}{c}{\normalsize{\textbf{Automated algorithms}}} \\ [0.5ex]
& \textbf{Division 1}  & \textbf{Division 2}  & \textbf{Division 3}  & \textbf{Division 4}\\ [0.5ex]
Number of annotators& 20  & 48  & 21  & 69\\ 
Average annotations & \multirow{2}{*}{18 (18$\star$)} & \multirow{2}{*}{39 (41$\star$)} & \multirow{2}{*}{15 (21$\star$)} & \multirow{2}{*}{54 (62$\star$)}  \\
per record&  & & & \\ [0.5ex]
RMSE score (ms)&6.65 (6.67$\star$)&16.36 (16.34$\star$)&17.46 (17.33$\star$)&16.36 (16.34$\star$)\\ [0.5ex]
Interquartile& \multirow{2}{*}{30.40} & \multirow{2}{*}{35.77} & \multirow{2}{*}{128.00} & \multirow{2}{*}{57.00}  \\ [0.5ex]
range of score (ms)&&&&\\ 
\hline\noalign{\smallskip}
\end{tabular}
}
\label{table:data description of 5-second segment} 
\\ [1ex]
\parbox{5.9in}{Note: The annotator/algorithm having the lowest RMSE over the 5-second segment was selected to represent the best score. The results with $\star$ were published in the Challenge for a 2-minute segment.}
\end{table}

\clearpage

\clearpage

\begin{table}[tp]
\caption{The parameters of the BCLA  and their values for modelling the 2006 PCinC data set.}
\centering  
\begin{center}
\resizebox{15cm}{!} {
\begin{subtable}
{
\centering
\begin{tabular}{>{\centering\arraybackslash}p{1.5cm}>{\centering\arraybackslash}p{5cm}>{\centering\arraybackslash}p{2.3cm}}
\hline\noalign{\smallskip}
\textbf{Symbol}& \textbf{Definition}  & \textbf{Value} \\  [0.5ex]
 {$k_b$} & {shape of Gamma distribution for $b$} & {3 \ddag}  \\
 {$\vartheta_b$} & {scale of Gamma distribution for $b$} & {0.0002 \ddag}  \\
 {$\mu_\phi$} & {mean of the bias distribution} & {10 \dag}  \\
 {$k_\alpha$} & {shape of Gamma distribution for $\alpha_\phi$} & {3 \dag}  \\
 {$\vartheta_\alpha$} & {scale of Gamma distribution for $\alpha_\phi$} & {0.0005 \dag}  \\
 {$k_\lambda$} & {shape of Gamma distribution for ${\bf\lambda}$} & {4*}  \\
 {$\vartheta_\lambda$} & {scale of Gamma distribution for ${\bf\lambda}$} & {0.003*}  \\
\hline\noalign{\smallskip}
\end{tabular}}
\end{subtable}%
}
\label{table:BCLA_parameters} 
\\[0.5ex]
\parbox{5.9in}{Note: $b$ is the precision parameter for the model of the ground truth. $\alpha_\phi$ is the precision parameter for the model of the bias. ${\bf\lambda}$ refers to annotators' precision values. The values with * are determined with the assumption that the annotations provided by the best performing algorithm is $\pm5$ms away from the reference. The values with $\dag$ are derived from \cite{Hughes2006, Couderc2011, Worster2014}. The values with $\ddag$ are derived from \cite{Malik2002, Goldenberg2006, CliffordBook2006}. }
 \end{center}
\end{table}

\clearpage

\begin{table}[tp]
\caption{The RMSEs and the MAEs of the inferred labels using different strategies in the simulated data set.}
\centering  
\begin{center}
\resizebox{15cm}{!} {
\begin{subtable}
{
\centering
\begin{tabular}{>{\centering\arraybackslash}p{1.5cm}>{\centering\arraybackslash}p{1.3cm}>{\centering\arraybackslash}p{1.3cm}>{\centering\arraybackslash}p{1.3cm}>{\centering\arraybackslash}p{1.3cm}>{\centering\arraybackslash}p{1.3cm}}
\hline\noalign{\smallskip}
&\textbf{Best Annotator}& \textbf{Median} & \textbf{Mean}  & \textbf{EM-R}  & \textbf{BCLA} \\  [0.5ex]
\textbf{RMSE} (ms) & {34.91$\pm$0.74}*  & {18.84$\pm$0.38}*  & {13.11$\pm$0.31}* & {14.21$\pm$0.36} & {\underline{6.44$\pm$0.34}*\dag}  \\
[0.5ex]
\textbf{MAE} (ms)& {30.15$\pm$0.72}* & {12.60$\pm$0.36}  & {11.26$\pm$0.30}* &{12.64$\pm$0.36} &  {\underline{5.14$\pm$0.30}}*\dag \\
 [0.5ex]
\hline\noalign{\smallskip}
\end{tabular}}
\end{subtable}%
}
\label{table:BCLA_sim_comparison} 
\\[0.5ex]
\parbox{5.9in}{Results significantly different from others ($p<0.0001$) as shown in $\dag{}$ for the BCLA model and * (columns 2 to 4, and 6 only) for the EM-R.}
 \end{center}
\end{table}

\clearpage

\clearpage

\begin{table}[tb]
\caption{The RMSEs and the MAEs of the inferred labels using different voting approaches in the 2006 PCinC data set.}
\centering  
\begin{center}
\resizebox{15cm}{!} {
\begin{subtable}
{
\centering
\begin{tabular}{>{\centering\arraybackslash}p{0.3cm}>{\centering\arraybackslash}p{1.3cm}>{\centering\arraybackslash}p{1.3cm}>{\centering\arraybackslash}p{1.3cm}>{\centering\arraybackslash}p{1.3cm}>{\centering\arraybackslash}p{1.3cm}>{\centering\arraybackslash}p{1.3cm}}
\hline\noalign{\smallskip}
\normalsize{\multirow{1}{*}{\textbf{}}} &    \multicolumn{5}{c}{\normalsize{\textbf{RMSE (ms)}}} \\ [0.5ex]
\hline \noalign{\smallskip}
\textbf{Div} &\textbf{Best Annotator}& \textbf{Median} & \textbf{Mean}  & \textbf{EM-R}  & \textbf{BCLA} \\  [0.5ex]
\textbf{2} & {15.43$\pm$0.73}*  &{15.29$\pm$0.58}  & {16.17$\pm$0.54}* & {15.05$\pm$0.49} &  \underline{{12.57$\pm$0.67}}*\dag \\  [0.8ex]
\textbf{3} & {17.25$\pm$2.33}* & {19.16$\pm$0.88}  & {30.46$\pm$1.57}* &{18.92$\pm$0.82} &   \underline{{13.90$\pm$0.84}}*\dag  \\  [0.8ex]
\textbf{4} & {15.37$\pm$2.13}* & {14.43$\pm$0.57}*  & {17.61$\pm$0.55}* &{14.76$\pm$0.52} &  \underline{{11.78$\pm$0.63}}*\dag  \\  [0.8ex]
\hline\noalign{\smallskip}
\normalsize{\multirow{1}{*}{\textbf{}}} & \multicolumn{5}{c}{\normalsize{\textbf{MAE (ms)}}}\\  [0.5ex]
\hline \noalign{\smallskip}   
\textbf{Div} &\textbf{Best Annotator}& \textbf{Median} & \textbf{Mean}  & \textbf{EM-R}  & \textbf{BCLA} \\  [0.5ex]
\textbf{2} & {10.85$\pm$0.58}*  & {11.76$\pm$0.42} & {12.61$\pm$0.43}* & {11.81$\pm$0.40} & \underline{{$9.29\pm$0.45}}*\dag\\  [0.8ex]
\textbf{3} & {11.61$\pm$3.03}* & {14.04$\pm$0.55}  & {22.89$\pm$0.96}* &{14.12$\pm$0.60} &  \underline{{$10.28\pm$0.67}}*\dag \\  [0.8ex]
\textbf{4} & {11.17$\pm$2.32}* & {11.21$\pm$0.40}*  & {14.16$\pm$0.43}* &{11.49$\pm$0.41} &  \underline{{8.56$\pm$0.42}}*\dag  \\  [0.8ex]
\hline\noalign{\smallskip}
\end{tabular}}
\end{subtable}%
}
\label{table:BCLA_qt_comparison} 
\\[0.5ex]
\parbox{5.9in}{Results significantly different from others ($p<0.0001$) as shown in $\dag{}$ for the BCLA model and * (columns 2 to 4, and 6 only) for the EM-R.}
 \end{center}
\end{table}

\clearpage

\end{document}